\pdfoutput=1

\documentclass[11pt]{article}

\usepackage{EMNLP2023}

\usepackage{times}
\usepackage{latexsym}
\usepackage{graphicx}

\usepackage[T1]{fontenc}

\usepackage[utf8]{inputenc}

\usepackage{microtype}

\usepackage{inconsolata}
\usepackage{graphicx}
\usepackage{amsmath}
\usepackage{booktabs}
\usepackage{multirow}
\usepackage{subfigure}
\usepackage{appendix}
\usepackage{tablefootnote}
\usepackage{relsize}
\usepackage{tablefootnote}
\usepackage{makecell}

\usepackage{xcolor}
\usepackage{amssymb}
\usepackage{pifont}

\usepackage{tcolorbox}
\usepackage{wrapfig}

\newcommand{\eat}[1]{}
\newcommand{\eg}{{\em e.g.,~}}     
\newcommand{\ie}{{\em i.e.,~}}      
\newcommand{\etc}{{\em etc. }}    

\newcommand{\tabincell}[2]{\begin{tabular}{@{}#1@{}}#2\end{tabular}}  
\newcommand{\cmark}{\ding{51}}%
\newcommand{\xmark}{\ding{55}}%

\makeatletter
\newcommand*\myfontsize{%
  \@setfontsize\myfontsize{8}{9}%
}
\makeatother
\usepackage{adjustbox}

\title{Improving Zero-shot Sentence Decontextualisation\\with Content Selection and Planning}

\author{
Zhenyun Deng,
Yulong Chen, 
Andreas Vlachos \\
Department of Computer Science and Technology, University of Cambridge \\
\{zd302, yc632, av308\}@cam.ac.uk 
\\
}

\begin{document}
\maketitle
\begin{abstract}
Extracting individual sentences from a document as evidence or reasoning steps is commonly done in many NLP tasks. However, extracted sentences often lack context necessary to make them understood, \eg coreference and background information. To this end, we propose a content selection and planning framework for zero-shot decontextualisation, which determines what content should be mentioned and in what order for a sentence to be understood out of context. Specifically, given a potentially ambiguous sentence and its context, we first segment it into basic semantically-independent units. We then identify potentially ambiguous units from the given sentence, and extract relevant units from the context based on their discourse relations.
Finally, we generate a content plan to rewrite the sentence by enriching each ambiguous unit with its relevant units. Experimental results demonstrate that our approach is competitive for sentence decontextualisation, producing sentences that exhibit better semantic integrity and discourse coherence, outperforming existing methods.
\end{abstract}


\section{Introduction} \label{section_1}
The extraction of sentences from documents is a common step in many NLP tasks 
including summarization~\cite{liu2019fine,zhong2020extractive}, 
fact checking~\cite{thorne2018fever,schlichtkrull2024averitec},
question answering~\cite{Yang0ZBCSM18,trivedi2022musique}, 
and passage retrieval~\cite{karpukhin2020dense,xiong2020answering}.
However, the interpretation of sentences often relies on contextual information which is lost when they are considered without it.
Sentence decontextualisation aims to address this issue by rewriting sentences to be understandable without context, while retaining their original meaning~\cite{choi2021decontextualization}.

Prior research on sentence decontextualisation has focused on using the paragraph where the ambiguous sentence is located \cite{choi2021decontextualization} or generating QA pairs related to the sentence as \textit{necessary context} to rewrite the sentence \cite{newman2023question,deng2024document}. However, these methods still leave some implicit discourse information in the sentence unresolved due to their inability to capture \textit{necessary context}, \eg unresolved coreference, missing discourse and background \cite{zhang2022extractive,schlichtkrull2024averitec}. Therefore, effective content selection and planning are crucial, as they can enhance the quality of the decontextualised sentences by pre-selecting relevant content and rewriting the sentence to include pre-selected content. 
However, how to determine the \textit{necessary context} related to the ambiguous sentence and incorporate it to generate an understandable and unambiguous sentence remains a challenge.


\begin{figure*}
	\begin{center}
		{\scalebox{0.64} {\includegraphics{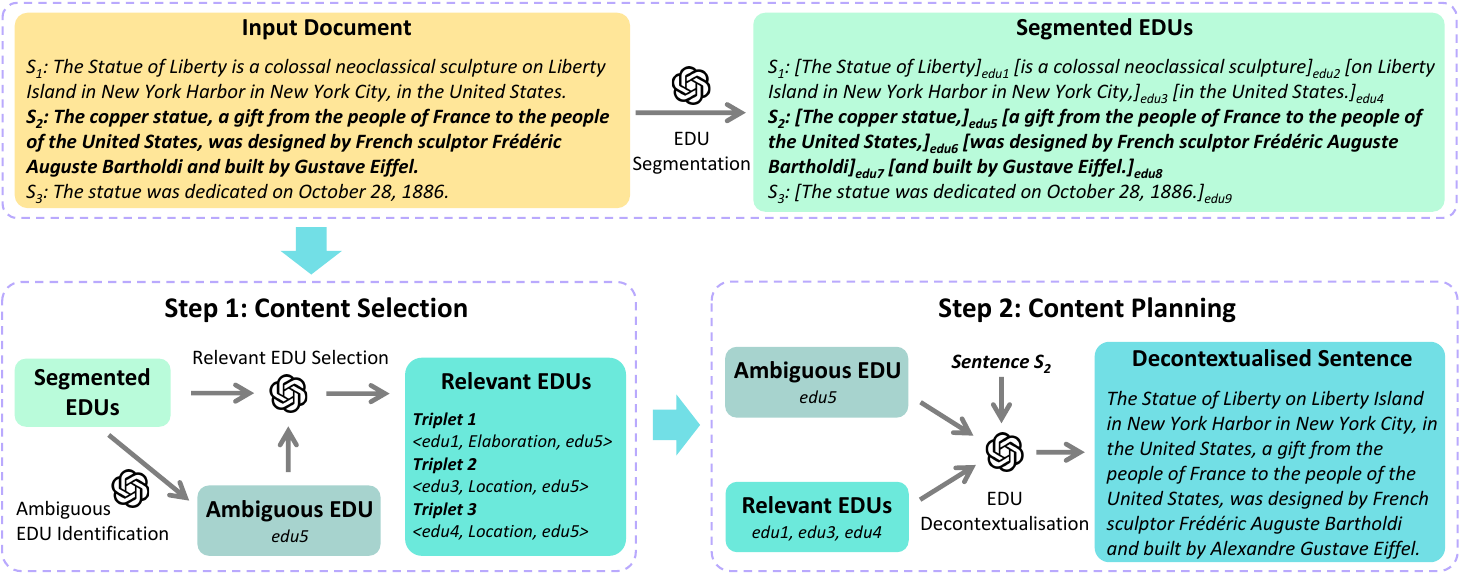}}}
		\caption{\footnotesize
		An overview of our proposed EDU-level content selection and planning (ECSP) framework for decontextualisation. The sentence to decontextualise is highlighted in bold. ECSP consists of two modules: $i)$ Content selection: identifies ambiguous EDUs in the sentence and selects EDUs that have discourse relations with the sentence as context required for decontextualisation; $ii)$ Content Planning: rewrites the sentence to be understood out of context by sequentially enriching each ambiguous EDU with its discourse-relevant EDUs.}
		\label{framework}
	\end{center}
	\vspace{-5mm}
\end{figure*}

Selecting the appropriate content granularity is essential for content selection, as it can help avoid introducing noise and redundant information. Entities or phrases are commonly considered in generation tasks \cite{cao2021controllable,fei2022cqg,xia2023improving}, but their discourse relations are unspecified and difficult to capture. 
Elementary Discourse Units (EDUs), widely used in discourse analysis, represent basic semantically independent spans within a sentence and are often employed to capture discourse relations~\cite{yang2018scidtb}.
For example, as shown in Figure~\ref{framework} (upper part), the sentence $S_2$ ``\textit{The copper statue, a gift ...}'' can be segmented into multiple EDUs, such as ``\textit{The copper statue}'' (edu5) and ``\textit{a gift ...}'' (edu6). 
Based on those fundamental semantic units, we can further identify discourse relations both within and across sentences, \eg the ``Elaboration'' relationship between ``\textit{The Statue of Liberty}'' (edu1) and edu5, and the ``Location'' relationship between ``\textit{on Liberty Island in New Your Harbor in New City / in the United States.}'' (edu3 / edu4) and edu5. Such EDUs can provide rich semantic and fine-grained discourse information, which is useful for downstream tasks such as abstractive summarisation~\cite{li-etal-2020-composing,delpisheh2024improving}.
However, how to effectively leverage them for decontextualisation remains a challenge.

\footnotetext[1]{\url{https://github.com/Tswings/ECSP}}

To this end, we propose an EDU-level Content Selection and Planning (ECSP) framework\protect\footnotemark[1], which determines what content should be selected and in what order, for decontextualisation. Specifically, $i$) to extract the \textit{necessary context}, we first segment the sentence and its context into EDUs, and then extract binary discourse relation pairs relevant to the sentence, where subordinate EDUs in relation pairs are identified as potentially ambiguous EDUs and their dominant EDUs are \textit{necessary context} to clarify them~\cite{yang2018scidtb}; $ii$) to improve the quality of the decontextualised sentences, we generate a content plan to rewrite the sentence to be understood without context by sequentially enriching each ambiguous EDU with its dominant EDUs, ensuring that each ambiguous EDU is clarified using its discourse-relevant content.

We evaluate ECSP on a benchmark dataset \cite{choi2021decontextualization} consisting of triplets containing (sentence, context, decontextualised sentence). ECSP outperforms two popular methods, SEGBOT \cite{li2018segbot} and NeuralSeg \cite{wang2018toward}, in the EDU segmentation task, as EDUs segmented by ECSP have better performance on semantic integrity and coherence. Additionally, unlike existing methods that solely generate decontextualised sentences, ECSP also identifies ambiguous EDUs within the sentence and provides relevant EDUs required for decontextualisation. In particular, ECSP achieves 87.5\% precision on identifying ambiguous EDUs and 83.98\% precision on selecting relevant EDUs. Furthermore, ECSP achieves the best scores on the decontextualiation task across multiple metrics, including SARI, BERTScore, ChrF, RougeL, BLEU and METEOR, outperforming all baselines. This is further supported by results on a fact-checking claim extraction dataset \cite{deng2024document}, as ECSP achieves a better ChrF score of 28.3 against gold decontextualised claims. When evaluated on the multi-hop QA dataset \cite{yang2018hotpotqa} for its potential in multi-hop reasoning and evidence retrieval, the QA model using our decontextualised evidence achieves a 1.58 improvement in F1 score for answer prediction and a 0.44 improvement in F1 score for evidence retrieval.

\section{Methodology}
Given a sentence $s$ and its context $C$, the task of decontextualisation is to rewrite $s$ to be understood without context by enriching it with $C$. ECSP is a system that not only returns the decontextualised sentence $s'$ but also identifies ambiguous EDUs within $s$ and relevant EDUs used to clarify them. As shown in Figure~\ref{framework}, ECSP consists of two main modules: $i)$ Content Selection ($\S 2.1$), identifies ambiguous EDUs in the sentence, and selects their relevant EDUs from context as necessary context; $ii)$ Content Planning ($\S 2.2$), generates a content plan to rewrite the sentence to be understood without context by sequentially enriching each ambiguous EDU with its relevant EDUs.

\subsection{Content Selection} \label{section_3_1}
The content selection step aims to determine which pieces of information from the context should be selected for decontextualisation. To this end, we first segment the sentence $s$ and its context $C$ into a sequence of EDUs; next, we extract all binary discourse relation pairs $\mathbf P$ between $s$ and $C$; $\mathbf P$ is a set of triples, each of which can be represented as ($\rm EDU_{dom}$, $\rm r$, $\rm EDU_{sub}$), where $\rm EDU_{dom}$ is the dominant EDU, $\rm EDU_{sub}$ is the subordinate EDU and $\rm r$ represents the relation between $\rm EDU_{dom}$ and $\rm EDU_{sub}$; then, we identify the subordinate EDUs in triples as potentially ambiguous EDUs, since their meanings are often unclear without their dominant EDUs, whereas dominant EDUs can typically be understood on their own~\cite{yang2018scidtb}.
finally, for each ambiguous EDU, we take its dominant EDUs as relevant EDUs to clarify it, \ie necessary context required for decontextualisation.

\paragraph{EDU Segmentation.}
EDU segmentation is a fundamental and important step in discourse analysis, aiming to segment texts into a sequence of EDUs. As illustrated in Figure~\ref{framework}, given a sentence $s$ and its context $C=\{s_1, ..., s_n\}$ as the inputs, we obtain their segmented EDUs by directly prompting the Large Language Model ($\phi$), respectively:
\begin{eqnarray}
\begin{array}{l}
\begin{aligned}
& \mathrm{EDU}_{s} = \phi(Prompt_{seg}(s)),    \\
& \mathrm{EDU}_{C} = \phi(Prompt_{seg}(C)),
\end{aligned}
\end{array}
\label{eq3_1_1}
\end{eqnarray}
where $Prompt_{seg}$ is a natural language instruction that guides the LLM to segment the sentence into EDUs. $\mathrm{EDU}_{s}$ and $\mathrm{EDU}_{C}$ denote the sets of EDUs in ${s}$ and ${C}$, respectively. 

\paragraph{EDU Selection.} Following EDU segmentation, the next step is to identify potentially ambiguous EDUs from $\mathrm{EDU}_{s}$ and their relevant EDUs from $\mathrm{EDU}_{C}$.  Discourse Dependency Parsing (DDP) is the task of analysing the discourse structure of a document by determining the binary discourse dependencies between EDUs. As we previously mentioned, these relations are represented as ($\rm EDU_{dom}$, $\rm r$, $\rm EDU_{sub}$), where the dominant  $\rm EDU_{dom}$ is defined as the unit containing essential information in a discourse relation, while the subordinate $\rm EDU_{sub}$ is the unit providing supporting content. Similar to \citet{yang2018scidtb}, we follow \citet{carlson2001discourse} to use a deletion test to determine the dominant and subordinate EDUs: if removing one EDU in a binary discourse relation pair has an insignificant effect on the understanding of the other EDU, the removed EDU is treated as subordinate and the other as the dominant. Based on this, we identify the subordinate EDUs in $\mathrm{EDU}_{s}$ as potentially ambiguous EDUs in $s$, \ie those that rely heavily on additional content for understanding. Specifically, given a set of EDUs from the input sentence $s$, we identify the sub-set of EDUs that can be potentially ambiguous: 
\begin{eqnarray}
\begin{array}{l}
\begin{aligned}
\mathrm{A} = \phi(Prompt_{amb}(s, \mathrm{EDU}_{s})),
\end{aligned}
\end{array}
\label{eq3_1_2}
\end{eqnarray}
where $\mathrm{A}=\{\mathrm{A}_1, \mathrm{A}_2, \ldots, \mathrm{A}_i, \ldots \} \in \mathrm{EDU}_s$ is the set of ambiguous EDUs in $s$ , and $Prompt_{amb}$ is a natural language instruction designed for identifying ambiguous EDUs. 


Given the identified ambiguous EDUs, the next step is to select their dominant EDUs as the relevant content to clarify them. Since not all discourse relations contribute to decontextualisation, we mainly focus on those relations that can improve the clarity, consistency and coherence of ambiguous sentences.We list all discourse relations that can improve decontextualisation in Appendix~\ref{appendix_A1}. In particular, given an identified ambiguous EDU, we extract its relevant EDUs from $C$ by prompting the LLM with the ambiguous EDU and $\mathrm{EDU}_{C}$:
\begin{eqnarray}
\begin{array}{l}
\begin{aligned}
\mathrm{RelEDU}_i = \phi (Prompt_{sel}(\mathrm{A}_i, \mathrm{EDU}_C)),
\end{aligned}
\end{array}
\label{eq3_1_3}
\end{eqnarray}
where $Prompt_{sel}$ is a natural language instruction that guides the LLM to select relevant EDUs from $\mathrm{EDU}_C$, $\mathrm{RelEDU}_i$ is the set of relevant EDUs of $\mathrm{A}_i$. All EDUs related to $\mathrm{A}$ are denoted as $\mathrm{RelEDU}=$ $\{\mathrm{RelEDU}_1, \mathrm{RelEDU}_2, \ldots, \mathrm{RelEDU}_i, \ldots \}$.

\subsection{Content Planning} \label{section_3_2}
The content planning step aims to ensure that the selected content is presented in the generated text as intended. In this section, we generate an EDU-level content plan, \ie EDU decontextualisation, to rewrite the sentence to be understood without context by enriching it with the content obtained in the content selection step.

\paragraph{EDU Decontextualisation.}
Unlike the previous work~\cite{choi2021decontextualization}, we consider EDUs as the fundamental units of context required for decontextualisation. 
Since decontextualised sentences should remain as close as possible to their original form, we propose to rewrite the sentence by enriching each ambiguous EDU with its relevant EDUs. In addition to addressing the issues of coreference resolution, global scoping and bridge anaphora already handled in previous work \citep{choi2021decontextualization}, we further improve decontextualisation by enhancing the discourse structure of sentences. 
Given an ambiguous sentence $s$, the ambiguous EDUs $\mathrm{A}_{i}$ in $s$ and the corresponding relevant EDUs $\mathrm{RelEDU}_i$, we prompt the LLM to rewrite $s$ as follows:
\begin{eqnarray}
\begin{array}{l}
\begin{aligned}
{s}^{*} = \phi (Prompt_{dec}(s, \mathrm{A}, \mathrm{RelEDU})),
\end{aligned}
\end{array}
\label{eq3_2_1}
\end{eqnarray}
where $Prompt_{dec}$ is a natural language instruction for EDU decontexualisation and ${s}^*$ is the decontextualised sentence. The detailed prompt functions can be found in Appendix~\ref{appen:prompts}.

\begin{table*}
\small
	\centering
	\begin{tabular} {l |c |c |c |c |c |c} \toprule
            \tabincell{c}{Method}  &SARI &BERTScore &ChrF &RougeL &BLEU &METEOR \\  \midrule 
            \multicolumn{7}{c}{\textit{Fully-supervised}} \\
            \midrule
            Coreference Model~\cite{joshi2020spanbert}     &0.4116  &0.9327 &0.7703 &0.7428 &0.5644 &0.7907 \\
            T5-base \cite{choi2021decontextualization}        &0.4823  &0.9410 &0.8188 &0.7831 &0.6497 &0.8306 \\
            T5-3B~\cite{choi2021decontextualization}     &0.5183 &0.9535 &0.8237 &0.8262 &0.6707 &0.8484 \\ 
            T5-11B~\cite{choi2021decontextualization}            &\colorbox{lightgray}{0.5215}  &\colorbox{lightgray}{0.9582} &\colorbox{lightgray}{0.8268} &\colorbox{lightgray}{0.8309} &\colorbox{lightgray}{0.6763} &\colorbox{lightgray}{0.8511} \\  
            \midrule 
            \multicolumn{7}{c}{\textit{Zero-shot}} \\
            \midrule
            QADECONTEXT~\cite{newman2023question}     &0.4312  &0.9361 &0.7724 &0.7583  &0.5727 &0.7906 \\ 
            DCE~\cite{deng2024document}     &0.4422  &0.9348 &0.7802 &0.7561  &0.5817 &0.7921 \\ 
		Vanilla Prompt (\textit{Llama-3.1-8B})   &0.3597  &0.9281 &0.7436 &0.6299 &0.5159 &0.7715 \\
        Vanilla Prompt (\textit{Gemini-1.5-flash})   &0.3624  &0.9317 &0.7462 &0.6611 &0.5272 &0.7762 \\ 
        Vanilla Prompt (\textit{GPT-4o})   &0.3732  &0.9309 &0.7451 &0.6531 &0.5231 &0.7786 \\ \midrule 
		ECSP (\textit{Llama-3.1-8B})      &0.4772  &0.9386 &0.8168  &0.7492 &0.6554 &0.8301 \\
            ECSP (\textit{Gemini-1.5-flash})  &0.4858  &\colorbox{lightgray}{\bf 0.9450$^*$} &\colorbox{lightgray}{\bf 0.8204$^*$} &\colorbox{lightgray}{\bf 0.8059$^*$} &\colorbox{lightgray}{\bf 0.6611$^*$} &0.8312 \\
            ECSP (\textit{GPT-4o})        &\colorbox{lightgray}{\bf 0.4993$^*$}  &0.9413 &0.8193 &0.7897$^*$ &0.6581$^*$ &\colorbox{lightgray}{\bf 0.8331$^*$} \\
  \bottomrule
	\end{tabular}
	\caption{Overall performance of different decontextualisation methods on the benchmark dataset. In the zero-shot setting, ECSP achieves the best results, with the best scores per metric marked in boldface. Among the fully-supervised and zero-shot settings, the best scores are marked in gray. Statistical significance over the T5-base model computed with the $t$-test is indicated with * ($p$ < 0.05).}
	\label{tab2}
    \vspace{-3mm}
\end{table*}

\section{Experimental Setup}
\subsection{Dataset and Metrics} \label{section_4_1}
\paragraph{Dataset.} 
We evaluate our proposed method using the dataset from \cite{choi2021decontextualization},
a widely used benchmark for sentence decontextualisation, which consists of the triplets ($sentence$, $context$, $decontextualised \ sentence$). The $sentence$ is a single sentence from Wikipedia; the $context$ is the paragraph in which the $sentence$ is located; the $decontextualised \ sentence$ is the decontextualised form of the $sentence$. Appendix~\ref{appen:dataset} shows the details of the dataset statistics. 

The task of this work is to rewrite the $sentence$ based on the given $context$, making it understandable without $context$, while preserving its original meaning. We adopt six metrics to evaluate the performance of our method on the decontextualisation task: SARI~\cite{xu2016optimizing}, BERTScore~\cite{zhang2019bertscore},  ChrF~\cite{popovic2015chrf}, RougeL~\cite{lin2004rouge}, BLEU~\cite{papineni2002bleu} and METEOR~\cite{lavie-agarwal-2007-meteor}. Details of metrics can be found in Appendix~\ref{appen:metric}.


\subsection{Baselines} \label{section_4_2}
We compare our method with both supervised and unsupervised baselines. 
Coreference Model \cite{joshi2020spanbert} is a fully-supervised model based on SpanBERT \cite{joshi2020spanbert} that rewrites the sentence by replacing unresolved coreferences in the sentence. 
T5-base, T5-3B and T5-11B \cite{choi2021decontextualization} are fully-supervised models that use T5 models to rewrite the sentence based on the paragraph where the sentence is located. 
DCE \cite{deng2024document} is a method consisting of multiple pre-trained models that do not require fine-tuning, which rewrites the claim to be understood out of context by enriching it with the generated QA pairs. QADECONTEXT \cite{newman2023question} is a zero-shot method that uses LLMs to generate QA pairs related to the ambiguous sentence, and then prompts LLMs with these QA pairs to rewrite the sentence. Vanilla Prompt \cite{brown2020language} is a standard prompting method.

\subsection{Implementation Details}
For content selection, given a sentence and its context as input, we directly prompt the baseline LLMs with them to select relevant contents. Each experiment in the content selection module is run with 10 demonstration instances, \eg EDU segmentation, ambiguous EDU identification and EDU selection. For content planning, \ie EDU decontextualisation, we do not use decontextualised instances for in-context learning. We list all the prompts used in Appendix~\ref{appen:prompts}. We use Huggingface library for the \textit{Llama-3.1-8B}; Gemini API for the \textit{Gemini-1.5-flash} model; and OpenAI API for the \textit{GPT-4o} model. We set the max output tokens to 512, temperature to 0 for all experiments.

\section{Results}
\subsection{Main Results}
Table \ref{tab2} summarises the main results on the test set of the benchmark dataset. 
First, ECSP outperforms zero-shot baselines by a significant margin, \eg QADECONTEXT and DCE, demonstrating that extracting relevant EDUs as pre-selected content is more effective than using generated relevant QA pairs. Furthermore, ECSP outperforms Vanilla Prompt on every metric, confirming that utilising content selection and planning for decontextualisation is effective in improving the quality of the decontextualised sentences. We also observe that \textit{GPT-4o} and \textit{Gemini-1.5-flash} outperform \textit{Llama-3.1-8B} across metrics, indicating that more powerful LLMs achieve superior performance. 

Notably, ECSP surpasses the fully-supervised coreference model, indicating that simply solving coreference problems is insufficient for achieving sentence decontextualisation, 
ECSP outperforms the T5-base model but underperforms the T5-3B and T5-11B models. However, it is noticeable that all supervised models are fine-tuned on 11K samples while our method is unsupervised. Moreover, ESCP can address challenges that other baselines cannot, such as identifying ambiguous units in the sentence and extracting the necessary content from the context required to clarify them.

\subsection{Analysis}\label{sec:analysis}
Our ECSP approach consists of four components: EDU Segmentation, Ambiguous EDU Identification, EDU selection and EDU Decontextualisation. Next, we conduct separate experiments to evaluate them and provide a detailed analysis on the sources of performance gain.

\begin{table}[t]
    \small
	\centering
	\begin{tabular} {l| c |c} \toprule
		\tabincell{c}{Method}  & Integrity  & Coherence \\ \midrule  
            NeuralSeg       &84\%  &90\%  \\
		  SEGBOT          &82\%  &84\%   \\  \midrule 
            \multicolumn{3}{l}{EDU Segmenation}        \\
            \ \ \ - \textit{Llama-3.1-8B}          &82\%  &88\%  \\
            \ \ \ - \textit{Gemini-1.5-flash}          &86\%  &94\%  \\
            \ \ \ - \textit{GPT-4o}          &86\%  &96\%  \\
  \bottomrule
	\end{tabular}
	\caption{Human Evaluation of EDU Segmentation on Semantic Integrity and Coherence.
    }
	\label{tab3}
    \vspace{-5mm}
\end{table}

\paragraph{Evaluation of EDU Segmentation.}
Since our ECSP uses EDUs as the fundamental unit for content selection and planning, their quality directly affects the performance of subsequent components. 
To evaluate the quality of our segmented EDUs, we compare our method against SEGBOT \cite{li2018segbot} and NeuralSeg \cite{wang2018toward}, two widely used EDU segmentation baselines.
In particular, we randomly selected 50 examples from the test set and recruited two graduate students to conduct a human evaluation of the quality of segmented EDUs based on the following two dimensions:
$i)$ \textit{Semantic Integrity} assesses whether the segmented EDUs retain their original meaning in the sentence; $ii)$ \textit{Coherence} assesses whether the segmented EDUs collectively preserve the coherence structure from the input document.
For each dimension, we ask human evaluators to give a binary score from \{0, 1\}, where 0 indicates the segmentation is flawed in that dimension, and 1 indicates it is satisfactory or correct. 
As shown in Table~\ref{tab3}, we observe that EDUs segmented by \textit{GPT-4o} preserve semantic integrity in 86\% of the samples and coherence in 96\%, both outperforming NeuralSeg and SEGBOT, even without fine-tuning. 
This indicates that our EDU segmentation method can better preserve discourse structure and relationships between EDUs, resulting in clearer and coherent EDUs. Examples of evaluations on the segmented EDUs can be found in Appendix~\ref{appen:EDU_seg_examples}.

\begin{table}[t]
    \small
	\centering
	\begin{tabular} {l| c |c} \toprule
		\tabincell{c}{Method}  & \#Ambig.  & Precision \\  \midrule  
		  Coreference      &2.47  &65.28\%  \\  \midrule 
          \multicolumn{3}{l}{\textit{Ambiguous EDU Identification}} \\
		\ \ \ - \textit{Llama-3.1-8B}            &1.85  &81.34\%  \\
            \ \ \ - \textit{Gemini-1.5-flash}        &1.10  &85.54\%  \\
            \ \ \ - \textit{GPT-4o}                  &0.94  &87.50\%  \\
  \bottomrule
	\end{tabular}
	\caption{Results of different methods for ambiguous EDU identification. \#Ambig. denotes the average number of ambiguous EDUs identified by different methods. 
    }
	\label{tab4}
    \vspace{-3mm}
\end{table}

\begin{table*}
\small
	\centering
	\begin{tabular} {c |c |c |c |c |c |c |c} \toprule
            \tabincell{c}{Context} &LLM  &SARI &BERTScore &ChrF &RougeL &BLEU &METEOR \\  \midrule 
		\multirow{3}{*}{\textit{Original Context}} &\textit{Llama-3.1-8B} &0.4468 &0.9311 &0.7862 &0.7281 &0.6283 &0.8043 \\ 
            &\textit{Gemini-1.5-flash}  &0.4807 &0.9221 &0.8063 &0.7674 &0.6505 &0.8187\\
            &\textit{GPT-4o}  &0.4853 &0.9143 &0.8041 &0.7644 &0.6412 &0.8134 \\  
            \midrule 
            \multirow{3}{*}{\textit{\makecell{Selected EDUs \\ (w/o content planning)}}}
		&\textit{Llama-3.1-8B} &0.4659 &0.9212 &0.7996 & 0.7304 &0.6494 &0.8173 \\ 
            &\textit{Gemini-1.5-flash} &0.4836 &0.9375 &0.8148 &0.7857 &0.6522 &0.8194  \\
            &\textit{GPT-4o} &0.4882 &0.9363 &0.8113 &0.7719 &0.6475 &0.8218   \\ \midrule 
            \multirow{3}{*}{\textit{\makecell{Selected EDUs \\ (w content planning)}}}
		&\textit{Llama-3.1-8B}      &0.4772  &0.9386 &0.8168  &0.7492 &0.6554 &0.8301 \\ 
            &\textit{Gemini-1.5-flash}  &0.4858  &0.9450 &0.8204 &0.8059 &0.6611 &0.8312 \\
            &\textit{GPT-4o}        &0.4993  &0.9413 &0.8193 &0.7897 &0.6581 &0.8331 \\
  \bottomrule
	\end{tabular}
	\caption{Results of EDU Decontextualisation under different context settings. \textit{Orignal Context} and \textit{Selected EDUs} denote rewriting ambiguous EDUs using the original context or selected EDUs, respectively.  \textit{w content planning} and \textit{w/o content planning} denote rewriting ambiguous EDUs with or without content planning, respectively.
    }
	\label{tab7}
        \vspace{-3mm}
\end{table*}

\paragraph{Evaluation of Ambiguous EDU Identification.}
To evaluate the quality of the identified ambiguous EDUs, we compare our method with the coreference model~\citep{joshi2020spanbert}. 
Specifically, if an identified EDU or entity contains text spans that require rewriting, we consider it successfully identified. 
We report both the overall precision and the average number of identified EDUs/entities. As shown in Table~\ref{tab4}, our method effectively identifies the majority of ambiguous text spans, with \textit{GPT-4o}, \textit{Gemini-1.5-flash}, and \textit{Llama-3.1-8B} achieving identification precision of 87.50\%, 85.54\%, and 81.34\%, respectively.
It shows our method greatly outperforms the coreference model with a precision of 65.28\%. 
Furthermore, we observe that the coreference model tends to identify a larger number of spans (2.47 on average) compared to our methods. 
However, its lower precision can introduce redundant information, adding extraneous details to entities that are already unambiguous.
In comparison, our method ensures that identified spans are both relevant and necessary for subsequent rewriting.

\begin{table}[t]
    \small
	\centering
	\begin{tabular} {l| c | c} \toprule
		\tabincell{c}{Method}  & avg.context  & Precision \\  \midrule  
            Gold Context                  &137    &86.48\%  \\ \midrule 
            \multicolumn{3}{l}{\textit{EDU Selection}}        \\
            \ \ \ - \textit{Llama-3.1-8B}           &35    &80.52\%  \\
		\ \ \ - \textit{Gemini-1.5-flash}       &21    &82.90\%  \\
            \ \ \ - \textit{GPT-4o}                 &23    &83.98\%  \\
  \bottomrule
	\end{tabular}
	\caption{Results of EDU Selection. avg.context denotes the average length of context used for decontextualisation. Precision denotes the precision of different methods in selecting the necessary context.}
	\label{tab5}
        \vspace{-3mm}
\end{table}

\paragraph{Evaluation of EDU Selection}
As discussed in Section~\ref{section_3_1}, selecting relevant EDUs and incorporating them into the rewritten sentence can improve the quality of the output sentence. To verify the effect of EDU selection, we calculate the precision of selected EDUs by measuring whether selected EDUs contain context used for decontextualisation. Table~\ref{tab5} shows the performance of our EDU selection model on the test set of the benchmark dataset. We observe that the average length of the gold context is 137 words.
However, based on the statistics on the dataset, we found that decontextualisation requires only an average of 6 additional words from the context.
This further indicates the importance of effective content selection in improving decontextualization.
By incorporating content selection, our model significantly reduces the length of necessary context while preserving most of the relevant content. In particular, when using \textit{GPT-4o} and \textit{Gemini-1.5-flash}, our model selects an average of 21 and 23 words, respectively, while retaining 82.90\% and 83.98\% of relevant content (\ie necessary context required for decontextualisation). This leads to a substantial reduction in context length, improving both efficiency and effectiveness in decontextualisation.

\begin{table}[t]
    \small
	\centering
	\begin{tabular} {l| c |c } \toprule
		\tabincell{c}{Method}  & Feasible  & Unfeasible \\  \midrule  
            Coreference    &75\%  &25\% \\ \midrule
            \multicolumn{3}{l}{\textit{EDU Decontextualisation}} \\
            \ \ \ - \textit{Llama-3.1-8B}          &85\%  &15\%  \\
		\ \ \ - \textit{Gemini-1.5-flash}      &91\%  &9\%  \\
            \ \ \ - \textit{GPT-4o}                &93\%  &7\%  \\
  \bottomrule
	\end{tabular}
	\caption{The statistics of sentence decontextualisation. $\rm Feasible$ and $\rm Unfeasible$ denote the percentages of sentences that can/cannot be decontextualised, respectively.}
	\label{tab6}
    \vspace{-3mm}
\end{table}

\paragraph{Effect of EDU Decontextualisation}
During decontextualisation, we generate a content plan, EDU decontextualisation, to rewrite the sentence by sequentially enriching its ambiguous EDUs with their relevant EDUs. To verify the effect of content planning, we evaluate our method under three different settings: $i$) Original Context: rewrites each EDU using the full original context; $ii$) selected EDUs (w/o content planning): rewrites each EDU using selected EDUs without content planning; $iii$) selected EDUs (w content planning): rewrites each EDU using selected EDUs with content planning. The results in Table~\ref{tab7} show that rewriting ambiguous EDUs using their relevant EDUs is more effective than using the full original context, which further validates the importance of content selection. Additionally, we observe that rewriting ambiguous EDUs with content planning yields better results, indicating that sequentially rewriting each ambiguous EDU can maximise the likelihood that each ambiguous EDU is clarified, resulting in a clearer and more coherent sentence. Furthermore, sequentially rewriting provides greater flexibility in handling complex EDUs. Rewriting ambiguous EDUs without content planning may result in redundancy or the omission of key information, ultimately affecting the overall quality of the rewritten sentences. We describe the statistics of decontextualisation in Table~\ref{tab6}. The results show that our method decontextualises a higher proportion of ambiguous sentences compared to the coreference model. When using \textit{Gemini-1.5-flash}, the percentage of decontextualisation reaches 93\%. Examples of incorrect decontextualisation are presented in Appendix~\ref{appen:error_cases}.

\section{Impact of Decontextualisation on Downstream Tasks} 
\paragraph{Multi-hop Evidence Retrieval and Reasoning}
As described in Section~\ref{section_1}, isolated sentences often lack sufficient information, which may negatively affect downstream tasks when used as intermediate evidence.
To evaluate whether decontextualisation can improve multi-hop evidence retrieval and reasoning, we conduct experiments on HotpotQA dataset \cite{yang2018hotpotqa}. Under the same retriever, Beam Retrieval \cite{zhang2024end}, we decontextualise the gold first-hop evidence and then use it to retrieve the next-hop evidence. The results in Table \ref{tab8} show that ECSP achieves the best performance with a 0.44 improvement in F1 score over Beam Retrieval, which indicates that decontextualised evidence can better facilitate the retrieval of the next-hop evidence. For the multi-hop reasoning, we decontextualise gold evidence and then use them to answer multi-hop questions. Under the same QA method, ECSP outperforms all baselines on every metric, achieving an EM/F1 score of 74.54/87.28 with an improvement of 1.92/1.58 over Beam Retrieval, respectively. This further indicates that improving discourse coherence among evidence can lead to more complete evidence, resulting in more consistent and accurate multi-hop reasoning, thereby improving overall performance. Examples of decontextualisation for multi-hop reasoning can be found in \ref{appen:multihop}.

\begin{table}[t]
    \small
	\centering
	\begin{tabular} {l |l |l |l |l } \toprule
            \multirow{3}{*}{Method} &\multicolumn{2}{c|}{\makecell{Multi-hop \\ Retrieval}}  &\multicolumn{2}{c}{\makecell{Multi-hop \\ Reasoning}} \\
            \cmidrule(lr){2-3} \cmidrule(lr){4-5} &EM &F1 &EM &F1 \\ \midrule 
            Beam Retrieval   &97.63  &98.71 &72.62  &85.70 \\ \midrule
            Coreference      &97.69	 &98.73	&72.62	&85.88 \\ 
            T5-11B	         &98.32	 &99.04	&74.16	&87.11 \\ 
            DCE	             &97.89	 &98.54	&73.33	&86.26 \\ 
            Vanilla prompt	 &97.86	 &98.94	&72.92	&86.25 \\  \midrule 
            ECSP             &98.54$^*$  &99.15$^*$ &74.54$^*$  &87.28$^*$ \\ 
  \bottomrule
	\end{tabular}
	\caption{Results on Multi-hop retrieval and reasoning. Statistical significance over T5-11B computed with the $t$-test is indicated with * ($p$ < 0.05).}
	\label{tab8}
        \vspace{-5mm}
\end{table}

\paragraph{Claim Extraction} We compare ECSP with DCE on a claim extraction dataset containing decontextualised claim sentences \cite{deng2024document}. Results in Table \ref{tab9} show that our ECSP outperforms DCE, achieving a better ChrF/Sari/BERTScore score of 28.3/6.92/84.6, respectively, indicating that selecting EDUs related to the sentence as necessary context for decontextualisation is more effective than constructing QA pairs related to the sentence.

\begin{table}[h]
    \small
	\centering
	\begin{tabular} {l |c |c |c} \toprule
		\tabincell{c}{Method} & ChrF &SARI & BERTScore \\  \midrule 
            DCE         &26.4 &6.70 &83.8  \\
            ECSP        &28.3 &6.92 &84.6  \\ 
  \bottomrule
	\end{tabular}
	\caption{Results on Document-level claim extraction.}
	\label{tab9}
        \vspace{-5mm}
\end{table}

\section{Case Study}
\begin{figure*}
	\begin{center}
	\subfigure{\scalebox{0.51} {\includegraphics{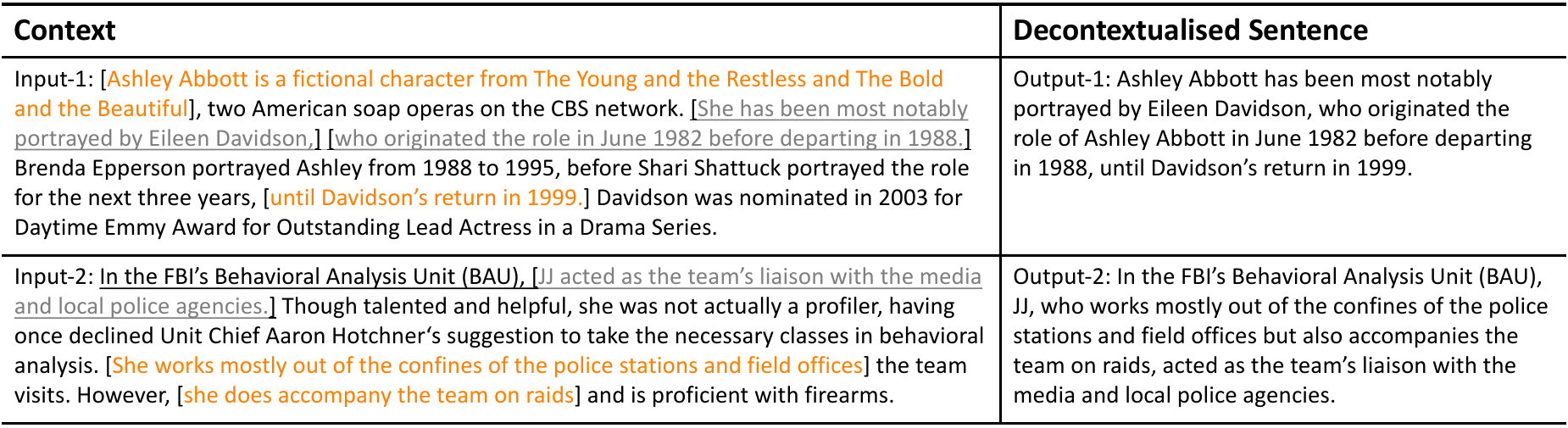}}}
        \vspace{-7mm}
        \caption{Case studies of our EDU decontextualisation. The sentences underlined are the ones to be \underline{decontextualised}. The text spans (\ie EDU) in gray are \textcolor{gray}{ambiguous EDUs} and in orange are \textcolor{orange}{relevant contextual EDUs}.
        }\label{tab10-1}
	\end{center}
        \vspace{-7mm}
\end{figure*}

We present two case examples in Figure~\ref{tab10-1}. Generally, the decontextualised sentences are grammatically fluent, consistent with the input sentences and their context, free from ambiguity, and easily understandable without the original context.
In particular, in the first decontextualised sentence (Output-1), we observe that the pronoun ``\textit{She}'' in the original sentence is replaced with the correct named entity, ``\textit{Ashley Abbott}''. Moreover, it enriches context with a time argument, ``\textit{until Davidson's return in 1999}''.
In the second case, to interpret the term “\textit{JJ},” the decontextualised sentence (Output-2) inserts an embedded clause (“\textit{who works ... but also accompanies ...}”) by combining and paraphrasing two individual sentences from context.
Both cases demonstrate the effectiveness of decontextualisation in improving clarity and coherence. We present more cases in Appendix~\ref{appen:case}.

\section{Related Work}
\paragraph{Content Selection and Planning}
Content selection and planning involve determining which pieces of information should be selected and in what order, to generate coherent text. Existing methods can be broadly divided into two categories: phrase-based and sentence-based content planning. Phrase-based methods extract key phrases from the given context and generate text based on extracted phrases. \citet{pan2020semantic} introduce a content selector to select question-worthy phrases from the semantic graph to generate questions. \citet{fei2022cqg} use graph attention networks to extract key entities in multi-hop reasoning chains, and then use a BERT-based decoder to ensure that these key entities appear in the generated question. Sentence-based methods focus on selecting key sentences to reduce the length of context.
Unlike the above methods, we choose EDU as the composition unit of the content selection because it provides richer semantic and fine-grained discourse information.


\paragraph{Elementary Discourse Unit}
Elementary discourse units are the smallest units of discourse and are often designed to capture the core information of a sentence. 
\citet{li2020composing} use an EDU selector to extract salient information and combine them together to generate a fluent summary. 
\citet{chen2021structure} propose a seq2seq model to improve abstractive conversation summarization models by constructing the EDU-based discourse graph and action graph.
In this work, we introduce an EDU identifier and an EDU selector to improve decontextualisation by identifying ambiguous EDUs in a sentence and their relevant EDUs.

\paragraph{Sentence Decontextualization}
Decontextualisation aims to rewrite a sentence to be understood out of context by enriching it with its context. Existing methods primarily rely on coreference resolution models or seq2seq generative models. \citet{joshi2020spanbert} mask contiguous random spans in the ambiguous sentence, and then predict the entire content of the masked spans to clarify the sentence. 
This method only solves ambiguous references in the sentence and does not introduce additional key information, such as background and temporal, that facilitate understanding the sentence without context.
\citet{choi2021decontextualization} use a T5-based method to rewrite the ambiguous sentence based on the paragraph where the sentence is located. \citet{mo2024controllable} propose a transformer-based sequence model that uses a soft-constraints mechanism to controllably rewrite polar questions and answers into decontextualised factual statements.
Although these methods introduce additional information to make the sentence clearer, they fail to capture discourse information that are important for sentence decontextualisation, such as cause-effect, condition and contrast. 
Without this information, decontextualised sentences may lose their original meaning and coherence, becoming ambiguous or potentially leading to misinterpretation. Unlike them, we introduce richer discourse information by identifying EDUs that have discourse relations with the ambiguous sentence, and then use them to rewrite the sentence to make it understandable out of context. 

\section{Conclusions}
This paper presented ECSP, an EDU-level content selection and planning framework for decontextualisation that rewrites the sentence to be understood out of context by enriching each ambiguous EDU with its relevant EDUs. We show that our method not only provides the decontextualised sentence but also identifies ambiguous EDUs and corresponding EDUs needed for clarification. Experimental results show that ECSP produces more coherent and comprehensible decontextualised sentences while achieving competitive performance in identifying ambiguous EDUs and relevant EDUs, highlighting its interpretability ability.
Future work looks at extending the capability of our method to more complex scenarios, including multimodal tasks.

\section*{Acknowledgements}
\setlength{\intextsep}{0pt}
\setlength{\columnsep}{8pt}
\begin{wrapfigure}{l}{0.44\columnwidth}
    \includegraphics[width=0.44\columnwidth]{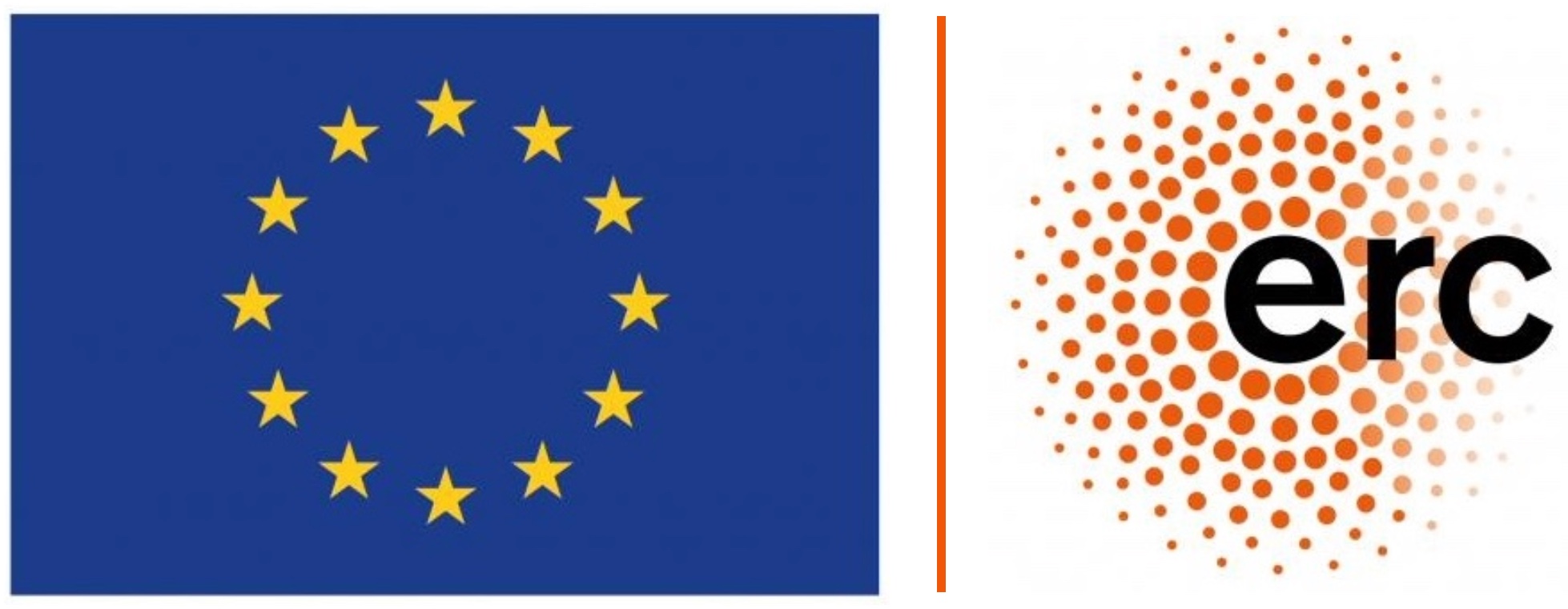} 
\end{wrapfigure}
This project has received funding from the European Research Council (ERC) under the European Union’s Horizon 2020 Research and Innovation programme grant AVeriTeC (Grant agreement No. 865958).  
We would like to thank the anonymous reviewers for their helpful questions and comments that helped us improve the paper.

\section*{Limitations}
While our method provides strong interpretability in identifying ambiguous units in the sentence and selecting relevant contents from context for sentence decontextualisation, it does not attempt to identify exact ambiguous text spans, as different text spans containing ambiguous units can be considered correct.
Instead, we focus on identifying ambiguous EDUs that cannot be clearly understood out of context. Additionally, our method relies on LLMs to segment texts into EDUs. While LLMs perform EDU segmentation well in most cases, improper segmentation can still impact the efficiency of decontextualisation, especially for texts requiring domain-specific knowledge. Moreover, our method achieves decontextualisation by rewriting ambiguous EDUs with their relevant EDUs; however, for different types of ambiguous EDUs, different decontextualisations may be considered correct, and more flexible content planning is worth further exploring.
Moreover, although our method is unsupervised, it relies on the strong capacity of LLMs. However, the experiments show that our method is universal across different LLMs, and it outperforms strong supervised methods.

\section*{Ethical Consideration}
We conducted human evaluation to measure the model performance on the EDU segmentation task (\autoref{sec:analysis}), with the help of two voluntary human evaluators. 
These two evaluators are doctoral students, who study in an English-speaking country and are specialised in NLP and discourse analysis.
During the evaluation, all system outputs were anonymised and presented to the evaluators in a randomised order. For each system output, the evaluators were asked to provide binary scores (\{0, 1\}) from two dimensions, \ie Semantic Integrity and Coherence, respectively.
We do not collect any personally sensitive information during the annotation.

\bibliography{custom}
\bibliographystyle{acl_natbib}


\appendix
\setcounter{table}{0}
\setcounter{figure}{0}
\setcounter{section}{0}
\setcounter{equation}{0}
\renewcommand{\thetable}{A\arabic{table}}
\renewcommand{\thefigure}{A\arabic{figure}}
\renewcommand{\thesection}{A\arabic{section}}
\renewcommand{\theequation}{A\arabic{equation}}

\onecolumn


\section{Dataset Statistics}\label{appen:dataset}
Details of the benchmark dataset~\cite{choi2021decontextualization} statistics.
\begin{table}[!h]
\small
	\centering
	\begin{tabular} {l |c |c |c} \toprule
            \tabincell{c}{Data}  &\#$\rm sample$  &$\rm avg. context$  &$\rm avg. sentence$ \\  \midrule 
            Train &11290  &132 &31.0    \\
 		Dev   &1945   &131 &31.2     \\
		  Test  &1945   &134 &31.5    \\
  \bottomrule
	\end{tabular}
	\caption{Descriptive statistics for the benchmark dataset. \#$\rm sample$ denotes the number of samples in this dataset, $\rm avg. context$ denotes the average length of context in words, $\rm avg. sentence$ denotes the average length of sentences in words.}
	\label{tab1}
\end{table}

\section{Evaluation Metric.}\label{appen:metric}
We use SARI \cite{xu2016optimizing}, ChrF \cite{popovic2015chrf} and BERTScore \cite{zhang2019bertscore}, which have been used in previous research \cite{choi2021decontextualization,deng2024document}, to evaluate the model performance. 
SARI is used to compare the sentence with the reference sentence by calculating the goodness of words that are added, deleted and kept. ChrF is used to compute the similarity between sentences using the character n-gram F-score. BERTScore is used to compute the semantic overlap between the sentence and the reference sentence. 
Furthermore, we also report performance on RougeL \cite{lin2004rouge}, BLEU \cite{papineni2002bleu}, METEOR \cite{lavie-agarwal-2007-meteor}, which are widely used for the text generation task. RougeL and BLEU are used to evaluate the recall and precision in $n$-gram matching between the reference and the generated text, respectively. METEOR is a comprehensive metric that evaluates partial matches between the reference and the generated text, and accounts for variations in word order and synonyms.

\section{Details of Prompts}\label{appen:prompts}
We list all the prompts used in our ECSP framework in the following subsections.
\subsection{EDU Segmentation Prompt}





\begin{quote}
    \textit{You will be given a sentence. Your task is to segment this sentence into Elementary Discourse Units (EDUs).} \\
    \\
    \textit{Generate the output as shown in the examples below.} \\
    \textit{------------------------------}     \\
    \textit{...}    \\
    \textit{Sentence: $\{s\}$; Output: $\{edu_1, edu_2, ..., edu_i, ..., \}$}\\
    \textit{...}    \\
    \textit{------------------------------}     \\
    \textit{Input:} \\
    \textit{Sentence: $\{\}$}   \\
    \textit{Output:}
\end{quote}
where $s$ is the sentence to be decontextualised, $edu_i$ is the $i$-th EDU of $s$.

\subsection{Ambiguous EDU Identification Prompt}
\begin{quote}
    \textit{You will be given a sentence and its EDUs. Your task is to extract ambiguous EDUs that rely heavily on context or have implicit references from the given EDUs.}  \\  \\
    \textit{Generate the output as shown in the examples below.}\\
    \textit{------------------------------}     \\
    \textit{...}    \\
    \textit{Sentence: $\{s\}$; EDUs: $\{edu_1, edu_2, ..., $ $edu_i, ..., \}$; Output: $\{ambedu_1, ..., ambedu_k, ..., \}$}\\
    \textit{...}    \\
    \textit{------------------------------}     \\
    \textit{Input:} \\
    \textit{Sentence: $\{\}$; EDUs: $\{\}$}   \\
    \textit{Output:}
\end{quote}
where $ambedu_k$ is the $k$-th ambiguous EDU in $s$.

\subsection{EDU Selection Prompt}
\begin{quote}
    \textit{You will be given a paragraph consisting of multiple sentences and their corresponding EDUs; an ambiguous sentence and its EDUs. Your task is to select EDUs from the paragraph that have discourse relations with the EDUs in the ambiguous sentence.}\\

    \textit{Generate the output as shown in the examples below.}\\
    \textit{------------------------------}     \\
    \textit{...}    \\
    \textit{Paragraph: $\{s_1, ..., s_j, ..., \}$; EDUs in Paragraph: $\{edu^1_1, edu^2_1, ..., edu^1_j, edu^2_j, ...,  \}$; \\ Sentence: $\{s_i\}$; Ambiguous EDUs in Sentence: $\{ambedu^1_i, ..., ambedu^k_i, ..., \}$; \\ Output: $\{reledu^1_{i1}, ..., reledu^m_{ik}, ..., \}$}\\
    \textit{...}    \\
    \textit{------------------------------}     \\
    \textit{Input:} \\
    \textit{Paragraph: $\{\}$; EDUs in Paragraph: $\{\}$; \\ Sentence: $\{\}$; EDUs in Sentence: $\{\}$;}   \\
    \textit{Output:}
\end{quote}
where $reledu^m_{ik}$ is $m$-th relevant EDU of $ambedu^k_i$.

\subsection{EDU Decontextualisation Prompt}
\begin{quote}
    \textit{You will be given a sentence and its ambiguous EDUs, and EDUs relevant to these ambiguous EDUs. Your task is to rewrite the ambiguous sentence to be understandable by enriching each ambiguous EDU with its relevant EDUs, which involves resolving ambiguities, determining references, and filling in implicit information. We prefer the rewritten sentence to be as close as possible to its original form.}\\
    \\
    \textit{------------------------------}     \\
    \textit{Input:} \\
    \textit{Sentence: $\{\}$; Ambiguous EDUs in Sentence: $\{\}$;  EDUs relevant to the sentence: $\{\}$;}  \\
    \textit{Output:}
\end{quote}

\subsection{Vanilla Decontextualisation Prompt}
\begin{quote}
    \textit{To rewrite the Sentence to be understandable out of Context, while retaining its original meaning. We prefer the rewritten sentence to be as close as possible to its original form.}       \\
    \\
    \textit{------------------------------}     \\
    \textit{Input:} \\
    \textit{Sentence: $\{\}$; Context: $\{\}$;}  \\
    \textit{Output:}
\end{quote}

\newpage
\section{Discourse Relation Category}\label{appen:discourse_relation}
Following previous work \cite{yang2018scidtb}, we list categories of discourse relations in Table \ref{appendix_A1}, where Decontext. Gain denotes the discourse relations that contribute to sentence decontextualisation. Since decontextualised sentences are preferred to be as close as possible to their original form~\cite{schlichtkrull2024averitec,deng2024document}, adding information from unrelated discourse relations may not improve sentence clarity, \eg, attribution, comparison, \etc Instead, it may be erroneously punished. Thus, we do not consider discourse relations that are not directly beneficial for decontextualisation.

\begin{table}[!h]
    \small
	\centering
	\begin{tabular} {l | c | c} \toprule
		\tabincell{c}{Coarse}  &Fine & Decontext. Gain  \\  \midrule
                Root            &Root  &{\xmark}            \\
                Attribution     &Attribution  &{\xmark}      \\ 
                Background      &General, Related & {\cmark}    \\
                Cause-effect    &Cause Result &{\cmark}            \\
                Comparison      &Comparison  &{\xmark}              \\
                Condition       &Condition   &{\cmark}              \\
                Contrast        &Contrast    &{\cmark}           \\
                Elaboration     &Addition, Definition &{\cmark}    \\
                Enablement      &Enablement &{\xmark}          \\
                Evaluation      &Evaluation  &{\xmark}         \\
                Explain         &Evidence, Reason &{\cmark}     \\
                Joint           &Joint, Coordination &{\xmark}   \\
                Manner-means   & Manner-means &{\xmark}       \\      
                Progression     &Progression &{\xmark}        \\
                Same-unit       &Same-unit &{\xmark}          \\
                Summary         &Summary  &{\xmark}           \\
                Temporal        &Temporal  &{\cmark}          \\
  \bottomrule
	\end{tabular}
	\caption{Categories of discourse relations.}
	\label{appendix_A1}
\end{table}




\section{Inference Time Statistics}\label{appen:infer_time}
Since ECSP involves additional steps compared to vanilla prompting, such as EDU segmentation, ambiguous EDU identification, EDU selection and EDU decontextualisation, which inevitably lead to increased computation time during inference. Thus, we conduct a comparative experiment on the dev set of the benchmark dataset to assess the efficiency of our ECSP compared and vanilla prompting. As shown in Table~\ref{tab:inf_time}, our ECSP requires 4 LLM calls per inference, resulting in longer inference time compared to Vanilla. However, this yields improved decontextualisation performance and provides intermediate outputs (\eg ambiguous units and relevant context) that enhance the interpretability of decontextualisation. This reflects a trade-off between the efficiency and performance of our method. We further report the average inference time for each module in ECSP in Table~\ref{tab:avg_time}.

\begin{table}[!h]
    \small
	\centering
	\begin{tabular} {l| c | c} \toprule
		\tabincell{c}{Method}  &\# calls   & avg.inference time \\  \midrule  
            DCE                    &0    &30.1298  \\
            Vanilla prompting      &1    &0.6482  \\
            QADECONTEXT            &3    &14.3735 \\ \midrule
            ECSP                   &4    &6.6426  \\
    \bottomrule
	\end{tabular}
	\caption{The comparison of efficiency between our ECSP and Vanilla prompting. \# $\rm calls$ denotes the number of large LLM calls per sample. $\rm avg.inference \ time$ denotes the average inference time per sample.} 
	\label{tab:inf_time}
\end{table}

\begin{table}[!h]
\small
	\centering
	\begin{tabular} {l |c |c |c |c |c } \toprule
            \tabincell{c}{Method}  &EDU Seg.  &Ambiguous EDU Iden. &EDU Sel. &EDU Decontext. & Total Avg. time  \\ \midrule 
            ECSP     &4.047  &0.638 &0.9851 &0.972 &6.6426 \\
  \bottomrule
	\end{tabular}
	\caption{The statistics of the average inference time for each component.}
	\label{tab:avg_time}
\end{table}

\newpage
\section{Reproducibility}\label{appen:reproduce}
To better reproduce our model, we also use the latest open-weight models, \textit{Llama-3.1-8B} and \textit{Llama-3.3-70B}, for sentence decontextualisation. The results in Table~\ref{tab:repro} indicate that more powerful \textit{Llama} models achieve better performance. We list all the prompts for decontextualisation in Appendix~\ref{appen:prompts}. 

\begin{table*}[!h] \label{appen:repro}
\small
	\centering
	\begin{tabular} {l |c |c |c |c |c |c} \toprule
            \tabincell{c}{Method}  &SARI &BERTScore &ChrF &RougeL &BLEU &METEOR \\  \midrule 
		Vanilla Prompt (\textit{Llama-3.1-8B})   &0.3597  &0.9281 &0.7436 &0.6299 &0.5159 &0.7715 \\
            Vanilla Prompt (\textit{Llama-3.3-70B})   &0.3668  &0.9312 &0.7456 &0.6592 &0.5254 &0.7769 \\
        \midrule 
	    ECSP (\textit{Llama-3.1-8B})    &0.4772  &0.9386  &0.8168  &0.7492 &0.6554 &0.8301 \\
            ECSP (\textit{Llama-3.3-70B})      &0.4935  &0.9427 &0.8202  &0.7964 &0.6598 &0.8315 \\
  \bottomrule
	\end{tabular}
	\caption{Results of different \textit{Llama} models on the benchmark dataset.}
	\label{tab:repro}
\end{table*}


\section{Case Study}\label{appen:case}
We provide some example sentences for case study. As shown in Table \ref{tab10}, to decontextualise the first example sentence, ECSP first identifies ambiguous EDUs in the sentence, \ie ``\textit{she has been most notably portrayed by Eileen Davidson}'' and ``\textit{who originated the role in June 1982 before departing in 1988.}''. Subsequently, ``\textit{Ashley Abbott is a fictional character from ...,}'' and ``\textit{until Davidson's return in 1999}'', as dominant EDUs of these two ambiguous EDUs, are selected as relevant EDUs for decontextualisation, where ``\textit{Ashley Abbott is a fictional character from ...,}'' provides the necessary background information that clarifies the ``she'' in the first ambiguous EDU (Background) and ``\textit{until Davidson's return in 1999}'' provides the temporal information for the second ambiguous EDU (Temporal). In the second example, ``She works mostly out of the confines of the police stations and field offices'' provides additional detail about JJ's work environment (Elaboration) and ``she does accompany the team on raids'' introduces an exception to JJ's office role (Contrast). In the third example, ``Bud Abbott stated that it was taken from an older routine called `Who's The Boss?''' provides the origin (Elaboration) and ``It was a big hit in 1937'' shows the effect of "hone the sketch" (Cause-effect). 
In the fourth example, ``On March 21, 2017, Apple announced an iPhone 7 ... '' provides more details about the ``it'' in the ambiguous sentence (Elaboration). 
In the fifth example, ``The law was introduced to the New Zealand Parliament as a private members bill
by Green Party Member of Parliament Sue Bradford in 2005,'' provides the specific content of the ``bill'' in the ambiguous EDU (Elaboration), and ``The bill was passed on its third reading on 16 May 2007'' provides the temporal information on when the bill was passed (Temporal).

\begin{table}[!h]
    \center  \myfontsize 
    \setlength{\tabcolsep}{0.5em}
    \begin{tabular}{p{9.5cm}p{6cm}}
        \toprule
            Context & Decontextualised sentence \\
        \midrule
            \textcolor{orange}{Bud Abbott stated that it was taken from an older routine called ``Who's The Boss?''}, a performance of which can be heard in an episode of the radio comedy program It Pays to Be Ignorant from the 1940s. \underline{\textcolor{gray}{After they formally teamed up}} \underline{\textcolor{gray}{in burlesque in 1936\textcolor{black}{,} he and Costello continued to hone the sketch}}. \textcolor{orange}{It was a big hit in 1937}, when they performed the routine in a touring vaudeville revue called ``Hollywood Bandwagon''.  &
            After Abbott and Costello formally teamed up in burlesque in 1936, they continued to hone the sketch, which was a big hit in 1937 and that Bud Abbott stated it was taken from an older routine called “Who’s The Boss?”. \\
            \cmidrule(lr){1-2}
            \textcolor{orange}{On March 21, 2017, Apple announced an iPhone 7 with a red color finish (and white front)}, as the part of its partnership with Product Red to highlight its AIDS fundraising campaign. \underline{\textcolor{gray}{It launched on March 24, 2017\textcolor{black}{,} but it was later}} \underline{\textcolor{gray}{discontinued after the announcement of the iPhone 8} and iPhone 8 Plus}.  &
            The iPhone 7 with a red color finish (and white front) launched on March 24, 2017, but it was later discontinued after the announcement of the iPhone 8 and iPhone 8 Plus.
            \\
            \cmidrule(lr){1-2}
            \textcolor{orange}{The law was introduced to the New Zealand Parliament as a private members bill by Green Party Member of Parliament Sue Bradford in 2005}, after being drawn from the ballot. It drew intense debate, both in Parliament and from the public. The bill was colloquially referred to by several of its opponents and newspapers as the ``anti-smacking bill''. \textcolor{orange}{The bill was passed on its third reading on 16 May 2007} by 113 votes to eight. \underline{The Governor-General of New Zealand granted the bill} \underline{Royal Assent on 21 May 2007, and \textcolor{gray}{the law came into effect on 21 June 2007}}.  &
            The Governor-General of New Zealand granted the bill, introduced to the New Zealand Parliament as a private members bill by Green Party Member of Parliament Sue Bradford in 2005 and passed on its third reading on 16 May 2007, Royal Assent on 21 May 2007, and the law came into effect on 21 June 2007. \\
        \bottomrule
    \end{tabular}
    \caption{Case studies of EDU-level decontextualisation. The sentences underlined are the ones to be decontextualised. The text spans (\ie EDU) in gray are ambiguous EDUs. The EDUs in orange are EDUs related to the ambiguous EDUs. }\label{tab10}
    \vspace{-3mm}
\end{table}

\newpage
\section{Examples of evaluation results for segmented EDUs} \label{appen:EDU_seg_examples}
\begin{table}[!h]
    \center  \myfontsize 
    \setlength{\tabcolsep}{0.5em}
    \begin{tabular}  {p{6.5cm}p{6.5cm}p{1cm}p{1cm}} \toprule
            Sentence & EDUs Segmented by ECSP & Integrity & Coherence \\
        \midrule
            The Padres traded him to the Boston Red Sox before entering the final year of his contract during the 2010-11 offseason and he was traded again to the Dodgers in August 2012.
            & [The Padres traded him to the Boston Red Sox] [before entering the final year of his contract during the 2010-11 offseason] [and he was traded again to the Dodgers in August 2012.] & \quad \ \ 1 & \quad \ \ 1      \\
            \cmidrule(lr){1-4}
            She is forced to choose between the Phantom and Raoul, but her compassion for the Phantom moves him to free them both and allow them to flee.
            & [She is forced to choose between the Phantom and Raoul,] [but her compassion for the Phantom moves him to free them both] [and allow them to flee.] & \quad \ \ 1 & \quad \ \ 1 \\
            \cmidrule(lr){1-4}
            It is anticipated that the building could be completed by 2026 -- the centenary of Gaudí's death.
            & [It is anticipated that the building could be completed by 2026] [-- the centenary of Gaudí's death.]  & \quad \ \ 1 & \quad \ \ 1 \\
            \cmidrule(lr){1-4}
            Soil salinity can be reduced by leaching soluble salts out of the soil with excess irrigation water.
            & [Soil salinity can be reduced by leaching soluble salts out of the soil with excess irrigation water.] & \quad \ \ 0 & \quad \ \ 0 \\
            \cmidrule(lr){1-4}
            At the end of each season, players receive rewards based on the highest league achieved during that season.
            & [At the end of each season,] [players receive rewards based on the highest league] [achieved during that season.] & \quad \ \ 1 & \quad \ \ 0 \\
            \cmidrule(lr){1-4}
            She has been most notably portrayed by Eileen Davidson, who originated the role in June 1982 before departing in 1988.
            & [She has been most notably portrayed by Eileen Davidson,] [who originated the role in June 1982 before departing in 1988.] & \quad \ \ 1 & \quad \ \ 1 \\
            \cmidrule(lr){1-4}
            In the FBI's Behavioral Analysis Unit (BAU), JJ acted as the team's liaison with the media and local police agencies.
            & [In the FBI's Behavioral Analysis Unit (BAU), JJ acted as the team's liaison with the media and local police agencies.] & \quad \ \ 0 & \quad \ \ 0 \\
            \cmidrule(lr){1-4}
            It launched on March 24, 2017, but it was later discontinued after the announcement of the iPhone 8 and iPhone 8 Plus.
            & [It launched on March 24, 2017,] [but it was later discontinued after the announcement of the iPhone 8] [and iPhone 8 Plus.] & \quad \ \ 0 & \quad \ \ 1 \\
        \bottomrule
    \end{tabular}
    \caption{Examples of semantic integrity and coherence evaluation.}\label{appen:edu_seg_evl}
    \vspace{-3mm}
\end{table}

\newpage
\section{Error Cases of Sentence Decontextualisation}\label{appen:error_cases}

\begin{table}[!h]
    \center  \myfontsize 
    \setlength{\tabcolsep}{0.5em}
    \begin{tabular}  {p{4.5cm}p{5cm}p{5.5cm}} \toprule
            Sentence & Reference Sentence & Sentence Decontextualised by ECSP \\ \midrule
            She is forced to choose between the Phantom and Raoul, but her compassion for the Phantom moves him to free them both and allow them to flee.
            & Christine Daaé is forced to choose between the Phantom of the Opera and her childhood sweetheart, Raoul, but her compassion for the Phantom moves him to free them both and allow them to flee. & Christine, a chorus girl, is forced to choose between the Phantom of the Opera, who becomes the object of obsession, passion and love for her, and Raoul, her childhood sweetheart, with whom she falls in love; but her compassion for the mysterious Phantom of the Opera moves him to free Christine and Raoul and allow them to flee.    \\
            \cmidrule(lr){1-3}
            
            Soil salinity can be reduced by leaching soluble salts out of the soil with excess irrigation water.
            & Soil salinity, the salt content in the soil (a mixture of organic matter, minerals, gases, liquids, and organisms), can be reduced by leaching soluble salts out of soil with excess irrigation water. & Soil salinity can be reduced by leaching soluble salts out of the soil with excess irrigation water. \\
            \cmidrule(lr){1-3}
            
            The series is generally praised for what is perceived as a sort of medieval realism.
            & The series "Game of Thrones" is generally praised for what is perceived as a sort of medieval realism. & The series is generally praised for its historical fiction approach, which is perceived as a sort of medieval realism. \\
            \cmidrule(lr){1-3}
            
            At the end of each season, players receive rewards based on the highest league achieved during that season.
            & At the end of each Clash Royale season, players receive rewards based on the highest league achieved during that season. & At the end of each season, players receive rewards based on the highest league achieved during that season in one of nine different leagues added to the game in March 2017. \\
            \cmidrule(lr){1-3}

            She has been most notably portrayed by Eileen Davidson, who originated the role in June 1982 before departing in 1988.
            & Ashley Abbott, a fictional character from The Young and the Restless and The Bold and the Beautiful, who has been most notably portrayed by Eileen Davidson, who originated the role in June 1982 before departing in 1988.
            & She has been most notably portrayed by Eileen Davidson, who originated the role of Ashley Abbott in June 1982 before departing in 1988,  until Davidson's return in 1999. \\
            \cmidrule(lr){1-3}

            In the FBI's Behavioral Analysis Unit (BAU), JJ acted as the team's liaison with the media and local police agencies.
            & In the FBI's Behavioral Analysis Unit (BAU) from the CBS crime drama Criminal Minds, Jennifer Jareau acted as the team's liaison with the media and local police agencies.
            & In the FBI's Behavioral Analysis Unit (BAU), JJ, who works mostly out of the confines of the police stations and field offices but also accompanies the team on raids, acted as the team's liaison with the media and local police agencies. \\
            \cmidrule(lr){1-3}

            It launched on March 24, 2017, but it was later discontinued after the announcement of the iPhone 8 and iPhone 8 Plus.
            & An iPhone 7 with a red color finish launched on March 24, 2017, but it was later discontinued after the announcement of the iPhone 8 and iPhone 8 Plus.
            & It launched on March 24, 2017, but the iPhone 7 with a red color finish (and white front) was later discontinued after the announcement of the iPhone 8 and iPhone 8 Plus. \\   
            \cmidrule(lr){1-3}

            To redeem himself, Talbot infuses himself with the gravity-manipulating substance gravitonium and kills alien warriors who are attacking S.H.I.E.L.D., becoming the MCU version of Graviton.
            & Glenn Talbot, a fictional character appearing in American comic books published by Marvel Comics, infuses himself with the gravity-manipulating substance gravitonium and kills alien warriors who are attacking S.H.I.E.L.D., becoming the MCU version of Graviton.
            & To redeem himself after his rescue and his brainwashing is briefly activated, revealing S.H.I.E.L.D.'s location to Hydra, and being rendered comatose in season 4's ``World's End'', Talbot infuses himself with the gravity-manipulating substance gravitonium and kills alien warriors who are attacking S.H.I.E.L.D., becoming the MCU version of Graviton. \\   
        \bottomrule
    \end{tabular}
    \caption{Error cases of sentence decontextualisation.}
    \vspace{-3mm}
\end{table}

\newpage
\section{Examples of Sentence Decontextualisation for Multi-hop Reasoning}\label{appen:multihop}

\begin{table}[!h]
    \center  \myfontsize 
    \setlength{\tabcolsep}{0.5em}
    \begin{tabular}  {p{4.5cm}p{5cm}p{5.5cm}} \toprule
            Question & Golden Evidence & Decontextualised Evidence \\ \midrule
            What government position was held by the woman who portrayed Corliss Archer in the film Kiss and Tell?
            & 1. As an adult, she was named United States ambassador to Ghana and to Czechoslovakia and also served as Chief of Protocol of the United States. 2. Kiss and Tell is a 1945 American comedy film starring then 17-year-old Shirley Temple as Corliss Archer.
            & 1. As an adult, Shirley Temple was named United States ambassador to Ghana and to Czechoslovakia and also served as Chief of Protocol of the United States.
            2. Kiss and Tell is a 1945 American comedy film starring then 17-year-old Shirley Temple as Corliss Archer.
            \\
            \cmidrule(lr){1-3}  
            The arena where the Lewiston Maineiacs played their home games can seat how many people?
            & 1. The Androscoggin Bank Colisée (formerly Central Maine Civic Center and Lewiston Colisee) is a 4,000 capacity (3,677 seated) multi-purpose arena, in Lewiston, Maine, that opened in 1958. 2. The team played its home games at the Androscoggin Bank Colisée.
            & 1. The Androscoggin Bank Colisée (formerly Central Maine Civic Center and Lewiston Colisee) is a 4,000 capacity (3,677 seated) multi-purpose arena, in Lewiston, Maine, that opened in 1958. 
            2. The Lewiston Maineiacs played its home games at the Androscoggin Bank Colisée.
            \\
            \cmidrule(lr){1-3}
            What is the name of the fight song of the university whose main campus is in Lawrence, Kansas and whose branch campuses are in the Kansas City metropolitan area?
            & 1. Kansas Song (We’re From Kansas) is a fight song of the University of Kansas.
            2. The main campus in Lawrence, one of the largest college towns in Kansas, is on Mount Oread, the highest elevation in Lawrence.
            3. Two branch campuses are in the Kansas City metropolitan area: the Edwards Campus in Overland Park, and the university's medical school and hospital in Kansas City.
            & 1. Kansas Song (We’re From Kansas) is a fight song of the University of Kansas.
            2. The main campus of the University of Kansas is in Lawrence, one of the largest college towns in Kansas, is on Mount Oread, the highest elevation in Lawrence.
            3. Two branch campuses of University of Kansas are in the Kansas City metropolitan area: the Edwards Campus in Overland Park, and the university's medical school and hospital in Kansas City.
            \\
            \cmidrule(lr){1-3}
            What screenwriter with credits for "Evolution" co-wrote a film starring Nicolas Cage and Téa Leoni?
            & 1. David Weissman is a screenwriter and director.
            2. His film credits include "The Family Man" (2000), "Evolution" (2001), and ""When in Rome"" (2010).
            3. The Family Man is a 2000 American romantic comedy-drama film directed by Brett Ratner, written by David Diamond and David Weissman, and starring Nicolas Cage and Téa Leoni.
            & 1. David Weissman is a screenwriter and director.
            2. David Weissman’s film credits include "The Family Man" (2000), "Evolution" (2001), and ""When in Rome"" (2010).
            3. The Family Man is a 2000 American romantic comedy-drama film directed by Brett Ratner, written by David Diamond and David Weissman, and starring Nicolas Cage and Téa Leoni.
            \\
            \cmidrule(lr){1-3}
            The football manager who recruited David Beckham managed Manchester United during what timeframe?
            & 1. Their triumph was made all the more remarkable by the fact that Alex Ferguson had sold experienced players Paul Ince, Mark Hughes and Andrei Kanchelskis before the start of the season, and not made any major signings.
            2. Instead, he had drafted in young players like Nicky Butt, David Beckham, Paul Scholes and the Neville brothers, Gary and Phil.
            3. Sir Alexander Chapman Ferguson, CBE (born 31 December 1941) is a Scottish former football manager and player who managed Manchester United from 1986 to 2013.
            & 1. Their triumph was made all the more remarkable by the fact that Alex Ferguson had sold experienced players Paul Ince, Mark Hughes and Andrei Kanchelskis before the start of the season, and not made any major signings.
            2. Instead, Alex Ferguson  had drafted in young players like Nicky Butt, David Beckham, Paul Scholes and the Neville brothers, Gary and Phil.
            3. Sir Alexander Chapman Ferguson, CBE (born 31 December 1941) is a Scottish former football manager and player who managed Manchester United from 1986 to 2013.
            \\
            \cmidrule(lr){1-3}
            Brown State Fishing Lake is in a country that has a population of how many inhabitants?
            & 1. Brown State Fishing Lake (sometimes also known as Brown State Fishing Lake And Wildlife Area) is a protected area in Brown County, Kansas in the United States.
            2. As of the 2010 census, the county population was 9,984.
            & 1. Brown State Fishing Lake (sometimes also known as Brown State Fishing Lake And Wildlife Area) is a protected area in Brown County, Kansas in the United States.
            2. As of the 2010 census, the Brown county population was 9,984. 
            \\
            \cmidrule(lr){1-3}
            Roger O. Egeberg was Assistant Secretary for Health and Scientific Affairs during the administration of a president that served during what years?
            & 1. Richard Milhous Nixon (January 9, 1913 – April 22, 1994) was the 37th President of the United States from 1969 until 1974, when he resigned from office, the only U.S. president to do so.
            2. His other roles included Assistant Secretary for Health and Scientific Affairs in the Department of Health, Education, and Welfare (now the United States Department of Health and Human Services) during the Nixon administration and Dean of the University.
            & 1. Richard Milhous Nixon (January 9, 1913 – April 22, 1994) was the 37th President of the United States from 1969 until 1974, when he resigned from office, the only U.S. president to do so.
            2. The other roles of Roger O. Egeberg included Assistant Secretary for Health and Scientific Affairs in the Department of Health, Education, and Welfare (now the United States Department of Health and Human Services) during the Nixon administration and Dean of the University. \\  
            \cmidrule(lr){1-3}
            This singer of A Rather Blustery Day also voiced what hedgehog?
            & 1. It was written by Robert \& Richard Sherman and sung by Jim Cummings as "Pooh".
            2. He is known for voicing the title character from "Darkwing Duck", Dr. Robotnik from "Sonic the Hedgehog", and Pete.
            & 1. "A Rather Blustery Day" is a whimsical song from the Walt Disney musical film featurette, "Winnie the Pooh and the Blustery Day".
            2. Jim Cummings is known for voicing the title character from "Darkwing Duck", Dr. Robotnik from "Sonic the Hedgehog", and Pete.
            \\
        \bottomrule
    \end{tabular}
    \caption{Sentence Decontextualisation for Multi-hop Reasoning.}
    \vspace{-3mm}
\end{table}










\end{document}